\documentclass[11pt,a4paper]{article}
\usepackage[hyperref]{acl2019}
\usepackage{times}
\usepackage{latexsym}
\usepackage{url}
\usepackage{bm}
\usepackage{amsmath,amssymb}
\usepackage[]{algorithm2e}
\SetAlCapSty{small}
\usepackage{stmaryrd}
\usepackage{graphicx}
\usepackage{subfigure}
\usepackage{multirow}
\usepackage{booktabs}
\usepackage{amsmath,amsfonts,amssymb,bbm,epsfig,bm}
\usepackage{physics}
\usepackage[font=small]{caption}
\usepackage{pifont}

\DeclareMathOperator{\E}{\mathbb{E}}

\DeclareMathOperator*{\argmin}{arg\,min}

\newcommand{\cmark}{\ding{51}}%
\newcommand{\xmark}{\ding{55}}%
\definecolor{tpgray}{gray}{0.90}

\aclfinalcopy 
\def\aclpaperid{1240} 

\title{Improving Open Information Extraction
\\via Iterative Rank-Aware Learning}

\author{Zhengbao Jiang, \quad Pengcheng Yin, \quad Graham Neubig \\
  Language Technologies Institute \\
  Carnegie Mellon University \\
  \texttt{\{zhengbaj, pcyin, gneubig\}@cs.cmu.edu} \\}

\date{}

\begin{document}
\maketitle
\begin{abstract}
Open information extraction (IE) is the task of extracting open-domain assertions from natural language sentences.
A key step in open IE is confidence modeling, ranking the extractions based on their estimated quality to adjust precision and recall of extracted assertions.
We found that the extraction likelihood, a confidence measure used by current supervised open IE systems, is not well calibrated when comparing the quality of assertions extracted from \emph{different} sentences.
We propose an additional binary classification loss to calibrate the likelihood to make it more globally comparable, and an iterative learning process, where extractions generated by the open IE model are incrementally included as training samples to help the model learn from trial and error.
Experiments on OIE2016 demonstrate the effectiveness of our method.\footnote{Code and data are available at \url{https://github.com/jzbjyb/oie_rank}}
\end{abstract}

\section{Introduction}

Open information extraction~(IE, \newcite{sekine2006demand, Banko:2007:OIE}) aims to extract open-domain assertions represented in the form of $n$-tuples (e.g., \textit{was born in; Barack Obama; Hawaii}) from natural language sentences (e.g., \textit{Barack Obama was born in Hawaii}).
Open IE started from rule-based~\citep{Fader:2011:Reverb} and syntax-driven systems~\citep{Mausam:2012:ollie, Corro:2013:ClausIE}, and recently has used neural networks for supervised learning~\citep{Stanovsky:2018:SupOIE, Cui:2018:NeuOIE, Sun:2018:LogOIE, Duh:2017:CLOIE, Jia:2018:SupOIE2}.

A key step in open IE is confidence modeling, which ranks a list of candidate extractions based on their estimated quality.
This is important for downstream tasks, which rely on trade-offs between the precision and recall of extracted assertions. 
For instance, an open IE-powered medical question answering (QA) system may require its assertions in higher precision (and consequently lower recall) than QA systems for other domains.
For supervised open IE systems, the confidence score of an assertion is typically computed based on its extraction likelihood given by the model~\citep{Stanovsky:2018:SupOIE, Sun:2018:LogOIE}.
However, we observe that this often yields sub-optimal ranking results, with incorrect extractions of one sentence having higher likelihood than correct extractions of another sentence.
We hypothesize this is due to the issue of a disconnect between training and test-time objectives.
Specifically, the system is trained solely to raise likelihood of gold-standard extractions, and during training the model is not aware of its test-time behavior of ranking a set of system-generated assertions across sentences that potentially include incorrect extractions.

\begin{figure}[tb]
\includegraphics[width=\columnwidth, clip, keepaspectratio]{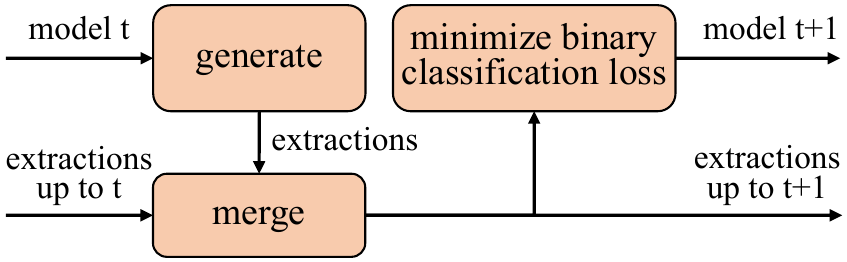}
\centering
\caption{Iterative rank-aware learning.}
\label{fig:arch}
\end{figure}

To calibrate open IE confidences and make them more globally comparable across different sentences, we propose an iterative rank-aware learning approach, as outlined in \autoref{fig:arch}.
Given extractions generated by the model as training samples, we use a binary classification loss to explicitly increase the confidences of correct extractions and decrease those of incorrect ones.
Without adding additional model components, this training paradigm naturally leads to a better open IE model, whose extractions can be further included as training samples.
We further propose an iterative learning procedure that gradually improves the model by incrementally adding extractions to the training data.
Experiments on the OIE2016 dataset \citep{Stanovsky:2016:OIE2016} indicate that our method significantly outperforms both neural and non-neural models.

\section{Neural Models for Open IE}\label{sec:oie}
We briefly revisit the formulation of open IE and the neural network model used in our paper.

\subsection{Problem Formulation}
Given sentence $\bm{s}=(w_1, w_2, ..., w_n)$, the goal of open IE is to extract assertions in the form of tuples $\bm{r}=(\bm{p}, \bm{a}_1, \bm{a}_2, ..., \bm{a}_m)$, composed of a single predicate and $m$ arguments.
Generally, these components in $\bm{r}$ need not to be contiguous, but to simplify the problem we assume they are contiguous spans of words from $\bm{s}$ and there is no overlap between them.

Methods to solve this problem have recently been formulated as sequence-to-sequence generation \citep{Cui:2018:NeuOIE, Sun:2018:LogOIE, Duh:2017:CLOIE} or sequence labeling \citep{Stanovsky:2018:SupOIE, Jia:2018:SupOIE2}.
We adopt the second formulation because it is simple and can take advantage of the fact that assertions only consist of words from the sentence.
Within this framework, an assertion $\bm{r}$ can be mapped to a unique BIO \citep{Stanovsky:2018:SupOIE} label sequence $\bm{y}$ by assigning $O$ to the words not contained in $\bm{r}$, $B_{p}$/$I_{p}$ to the words in $\bm{p}$, and $B_{a_i}$/$I_{a_i}$ to the words in $\bm{a}_i$ respectively, depending on whether the word is at the beginning or inside of the span.

The label prediction $\hat{\bm{y}}$ is made by the model given a sentence associated with a predicate of interest $(\bm{s}, v)$.
At test time, we first identify verbs in the sentence as candidate predicates.
Each sentence/predicate pair is fed to the model and extractions are generated from the label sequence.

\subsection{Model Architecture and Decoding}
Our training method in \autoref{sec:ours} could potentially be used with any probabilistic open IE model, since we make no assumptions about the model and only the likelihood of the extraction is required for iterative rank-aware learning.
As a concrete instantiation in our experiments, we use RnnOIE \citep{Stanovsky:2018:SupOIE, He:2017:DeepSRL}, a stacked BiLSTM with highway connections \citep{Zhang:2016:Highway, Srivastava:2015:TrainDeepNet} and recurrent dropout \citep{Gal:2016:RecDropout}.
Input of the model is the concatenation of word embedding and another embedding indicating whether this word is predicate:
\begin{equation*}
\bm{x}_t = [\bm{W}_{\text{emb}}(w_t), \bm{W}_{\text{mask}}(w_t = v)].
\end{equation*}
The probability of the label at each position is calculated independently using a softmax function:
\begin{equation*}
P(y_t|\bm{s}, v) \propto \text{exp}(\bm{W}_{\text{label}}\bm{h}_t + \bm{b}_{\text{label}}),
\end{equation*}
where $\bm{h}_t$ is the hidden state of the last layer.
At decoding time, we use the Viterbi algorithm to reject invalid label transitions \citep{He:2017:DeepSRL}, such as $B_{a_2}$ followed by $I_{a_1}$.%
\footnote{This formulation cannot easily handle coordination,
where multiple instances of an argument are extracted for a single predicate, so we use a heuristic of keeping only the first instance of an argument.}

We use average log probability of the label sequence \citep{Sun:2018:LogOIE} as its confidence:\footnote{The log probability is normalized by the length of the sentence to avoid bias towards short sentences. The original confidence score in RnnOIE is slightly different from ours. Empirically, we found them to perform similarly.}
\begin{equation}
c(\bm{s}, v, \hat{\bm{y}}) = \frac{\sum_{t=1}^{|\bm{s}|}{\log{P(\hat{y_t}|\bm{s}, v)}}}{|\bm{s}|}.
\label{eq:conf}
\end{equation}
The probability is trained with maximum likelihood estimation (MLE) of the gold extractions.
This formulation lacks an explicit concept of cross-sentence comparison, and thus incorrect extractions of one sentence could have higher confidence than correct extractions of another sentence.

\section{Iterative Rank-Aware Learning}\label{sec:ours}
In this section, we describe our proposed binary classification loss and iterative learning procedure.

\subsection{Binary Classification Loss}
\label{sec:ours:bin_loss}

To alleviate the problem of incomparable confidences across sentences, we propose a simple binary classification loss to calibrate confidences to be globally comparable.
Given a model $\theta^\prime$ trained with MLE, beam search is performed to generate assertions with the highest probabilities for each predicate.
Assertions are annotated as either positive or negative with respect to the gold standard, and are used as training samples to minimize the hinge loss:
\begin{equation}
\label{eq:loss}
\hspace{-1mm}\hat{\theta} = \underset{\theta}{\argmin}\hspace{-3mm}\underset{\substack{\bm{s} \in \mathcal{D}\\ v, \hat{\bm{y}} \in g_{\theta^\prime}(\bm{s})}}{\E}\hspace{-4mm}\max{(0,1-t \cdot
c_{\theta}(\bm{s}, v, \hat{\bm{y}}))},
\end{equation}
where $\mathcal{D}$ is the training sentence collection, $g_{\theta^\prime}$ represents the candidate generation process, and $t \in \{1,-1\}$ is the binary annotation. $c_{\theta}(\bm{s}, v, \hat{\bm{y}})$ is the confidence score calculated by average log probability of the label sequence.

The binary classification loss distinguishes positive extractions from negative ones generated across different sentences, potentially leading to a more reliable confidence measure and better ranking performance.

\subsection{Iterative Learning}
\label{sec:ours:iter_algo}
Compared to using external models for confidence modeling, an advantage of the proposed method is that the base model does not change: the binary classification loss just provides additional supervision.
Ideally, the resulting model after one-round of training becomes better not only at confidence modeling, but also at assertion generation, suggesting that extractions of higher quality can be added as training samples to continue this training process iteratively.
The resulting iterative learning procedure (\autoref{alg:iter}) incrementally includes extractions generated by the current model as training samples to optimize the binary classification loss to obtain a better model, and this procedure is continued until convergence.
\begin{algorithm}[t]
\small
\SetAlgoLined
\KwIn{training data $\mathcal{D}$, initial model $\theta^{(0)}$}
\KwOut{model after convergence $\theta$}
 $t \leftarrow 0$ \# iteration\\
 $\mathcal{E} \leftarrow \emptyset$ \# generated extractions\\
 \While{not converge}{
  $\mathcal{E} \leftarrow \mathcal{E} \cup \{(\bm{s}, v, \hat{\bm{y}})|v,\hat{\bm{y}} \in g_{\theta^{(t)}}(\bm{s}), \forall \bm{s} \in \mathcal{D}\}$\\
  $\theta^{(t+1)} \leftarrow \underset{\theta}{\argmin}\hspace{-3mm}\underset{(\bm{s}, v, \hat{\bm{y}})\in \mathcal{E}}{\E}\hspace{-3mm}\max{(0,1-t \cdot c_{\theta}(\bm{s}, v, \hat{\bm{y}}))}$\\
  $t \leftarrow t+1$\;
 }
 \caption{\small Iterative learning.}
 \label{alg:iter}
\end{algorithm}

\section{Experiments}\label{sec:exp}
\subsection{Experimental Settings}

\begin{table}[t]
\begin{center}
\small
\begin{tabular}{lrrr}
\toprule
 & \textbf{Train} & \textbf{Dev.} & \textbf{Test} \\
\midrule
\# sentence & 1\,688 & 560 & 641 \\
\# extraction & 3\,040 & 971 & 1\,729 \\
\bottomrule
\end{tabular}
\end{center}
\caption{Dataset statistics.}
\label{tab:data}
\end{table}

\paragraph{Dataset}
We use the OIE2016 dataset \citep{Stanovsky:2016:OIE2016} to evaluate our method, which only contains verbal predicates.
OIE2016 is automatically generated from the QA-SRL dataset \citep{He:2015:QASRL}, and to remove noise, we remove extractions without predicates, with less than two arguments, and with multiple instances of an argument.
The statistics of the resulting dataset are summarized in \autoref{tab:data}.

\paragraph{Evaluation Metrics}
We follow the evaluation metrics described by \newcite{Stanovsky:2016:OIE2016}: area under the precision-recall curve (AUC) and F1 score.
An extraction is judged as correct if the predicate and arguments include the syntactic head of the gold standard counterparts.%
\footnote{The absolute performance reported in our paper is much lower than the original paper because the authors use a more lenient lexical overlap metric in their released code: \url{https://github.com/gabrielStanovsky/oie-benchmark}.}

\paragraph{Baselines}
We compare our method with both competitive neural and non-neural models, including RnnOIE \citep{Stanovsky:2018:SupOIE}, OpenIE4,\footnote{\url{https://github.com/dair-iitd/OpenIE-standalone}} ClausIE \citep{Corro:2013:ClausIE}, and PropS \citep{Stanovsky:2016:PropS}.

\paragraph{Implementation Details}
Our implementation is based on AllenNLP \citep{Gardner:2018:AllenNLP} by adding binary classification loss function on the implementation of RnnOIE.\footnote{\url{https://allennlp.org/models\#open-information-extraction}}
The network consists of 4 BiLSTM layers (2 forward and 2 backward) with 64-dimensional hidden units.
ELMo \citep{Peters:2018:ELMo} is used to map words into contextualized embeddings, which are concatenated with a 100-dimensional predicate indicator embedding.
The recurrent dropout probability is set to 0.1.
Adadelta \citep{Zeiler:2012:Adadelta} with $\epsilon=10^{-6}$ and $\rho=0.95$ and mini-batches of size 80 are used to optimize the parameters.
Beam search size is 5.

\subsection{Evaluation Results}

\begin{figure}[tb]
  \subfigure{
    \includegraphics[width=0.9\columnwidth, clip, keepaspectratio]{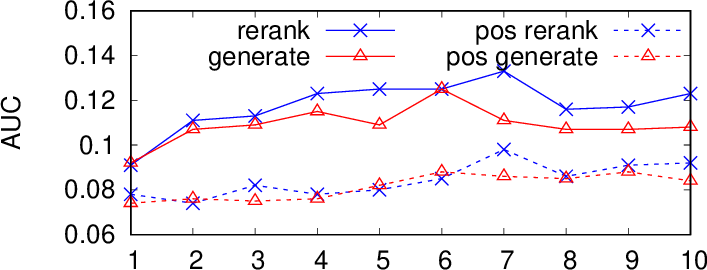}}
  \subfigure{
    \includegraphics[width=0.9\columnwidth, clip, keepaspectratio]{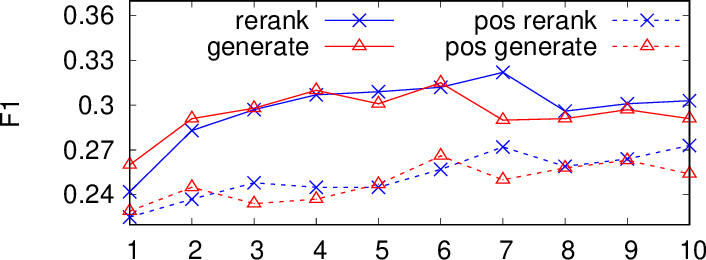}}
\caption{AUC and F1 at different iterations.}
\label{fig:iter}
\end{figure}

\begin{table*}[tb]
\small
\centering
\begin{tabular}{p{0.8\linewidth}|c c c}
\multirow{2}{*}{\textbf{sentence}} & \textbf{old} & \textbf{new} & \multirow{2}{*}{\textbf{label}} \\
 & \textbf{rank} & \textbf{rank} &  \\
\hline
\colorbox{blue!20}{\strut A CEN} \colorbox{red!20}{\strut forms} \colorbox{blue!20}{\strut an important but small part of a Local Strategic Partnership}. & 3 & 1 & \cmark \\
\hline 
\colorbox{blue!20}{\strut An animal} that cares for its young but shows no other sociality traits is said to \colorbox{red!20}{\strut be} ``\colorbox{blue!20}{\strut subsocial''}. & 2 & 2 & \xmark \\
\hline 
A casting director at the time told Scott that he had \colorbox{red!20}{\strut wished} \colorbox{blue!20}{\strut that he'd met him a week before}; he was casting for the ``G.I. Joe'' cartoon. & 1 & 3 & \xmark \\
\end{tabular}
\caption{Case study of reranking effectiveness. Red for predicate and blue for arguments.}
\label{tab:case_rerank}

\centering
\begin{tabular}{p{0.9\linewidth}|c}
\textbf{sentence} & \textbf{label} \\
\hline 
\colorbox{blue!20}{\strut A Democrat}, he \colorbox{red!20}{\strut became} \colorbox{blue!20}{\strut the youngest mayor in Pittsburgh's history} \colorbox{blue!20}{\strut in September 2006 at the age of 26}. & \cmark \\
\hline
\colorbox{blue!20}{\strut A motorcycle speedway long-track meeting}, one of the few held in the UK, \colorbox{red!20}{\strut was} staged at Ammanford. & \xmark \\
\end{tabular}
\caption{Case study of generation effectiveness. Red for predicate and blue for arguments.}
\label{tab:case_gen}
\end{table*}

\autoref{tab:exp_main} lists the evaluation results.
Our base model (RnnOIE,~\autoref{sec:oie}) performs better than non-neural systems, confirming the advantage of supervised training under the sequence labeling setting.
To test if the binary classification loss (E.q.~\ref{eq:loss}, \autoref{sec:ours}) could yield better-calibrated confidence, we perform one round of fine-tuning of the base model with the hinge loss (\textbf{$+$Binary loss} in~\autoref{tab:exp_main}).
We show both the results of using the confidence (E.q.~\ref{eq:conf}) of the fine-tuned model to rerank the extractions of the base model (\textbf{Rerank Only}), and the end-to-end performance of the fine-tuned model in assertion generation (\textbf{Generate}).
We found both settings lead to improved performance compared to the base model, which demonstrates that calibrating confidence using binary classification loss can improve the performance of both reranking and assertion generation.
Finally, our proposed iterative learning approach (\autoref{alg:iter}, \autoref{sec:ours}) significantly outperforms non-iterative settings.

We also investigate the performance of our iterative learning algorithm with respect to the number of iterations in \autoref{fig:iter}.
The model obtained at each iteration is used to both rerank the extractions generated by the previous model and generate new extractions.
We also report results of using only positive samples for optimization.
We observe the AUC and F1 of both reranking and generation increases simultaneously for the first 6 iterations and converges after that, which demonstrates the effectiveness of iterative training. The best performing iteration achieves AUC of 0.125 and F1 of 0.315, outperforming all the baselines by a large margin. 
Meanwhile, using both positive and negative samples consistently outperforms only using positive samples, which indicates the necessity of exposure to the errors made by the system.

\begin{table}[t]
\begin{center}
\small
\begin{tabular}{lrr}
\toprule 
\textbf{System} & \textbf{AUC} & \textbf{F1} \\
\midrule
\multicolumn{3}{c}{\it Non-neural Systems} \\
PropS \citep{Stanovsky:2016:PropS} & .006 & .065  \\
ClausIE \citep{Corro:2013:ClausIE} & .026 & .144  \\
OpenIE4 & .034 & .164  \\ \midrule
\multicolumn{3}{c}{\it Neural Systems} \\
Base Model (RnnOIE \newcite{Stanovsky:2018:SupOIE}) &  .050 & .204 \\
~$+$Binary~loss (\autoref{sec:ours:bin_loss}), Rerank Only &  .091 & .225 \\
~$+$Binary~loss (\autoref{sec:ours:bin_loss}), Generate &  .092 & .260 \\
~$+$Iterative Learning (\autoref{sec:ours:iter_algo}) &  {\bf .125} & {\bf .315} \\
\bottomrule
\end{tabular}
\end{center}
\caption{AUC and F1 on OIE2016.}
\label{tab:exp_main}
\end{table}

\paragraph{Case Study}
\autoref{tab:case_rerank} compares extractions from RnnOIE before and after reranking.
We can see the order is consistent with the annotation after reranking, showing the additional loss function's efficacy in calibrating the confidences; this is particularly common in extractions with long arguments.
\autoref{tab:case_gen} shows a positive extraction discovered after iterative training (first example), and a wrong extraction that disappears (second example), which shows that the model also becomes better at assertion generation.

\paragraph{Error Analysis}
Why is the performance still relatively low? We randomly sample 50 extractions generated at the best performing iteration and conduct an error analysis to answer this question. To count as a correct extraction, the number and order of the arguments should be exactly the same as the ground truth and syntactic heads must be included, which is challenging considering that the OIE2016 dataset has complex syntactic structures and multiple arguments per predicate.

\begin{table}[t]
\begin{center}
\small
\begin{tabular}{c c c}
\toprule 
\textbf{overgenerated} & \textbf{wrong} & \textbf{missing} \\
\textbf{predicate} & \textbf{argument} & \textbf{argument} \\
\midrule
41\% & 38\% & 21\% \\
\bottomrule
\end{tabular}
\end{center}
\caption{Proportions of three errors.}
\label{tab:err}
\end{table}

We classify the errors into three categories and summarize their proportions in \autoref{tab:err}. ``Overgenerated predicate'' is where predicates not included in ground truth are overgenerated, because all the verbs are used as candidate predicates. An effective mechanism should be designed to reject useless candidates. ``Wrong argument'' is where extracted arguments do not coincide with ground truth, which is mainly caused by merging multiple arguments in ground truth into one. ``Missing argument'' is where the model fails to recognize arguments. These two errors usually happen when the structure of the sentence is complicated and coreference is involved. More linguistic information should be introduced to solve these problems.

\section{Conclusion}
We propose a binary classification loss function to calibrate confidences in open IE. Iteratively optimizing the loss function enables the model to incrementally learn from trial and error, yielding substantial improvement. An error analysis is performed to shed light on possible future directions.

\section*{Acknowledgements}

This work was supported in part by gifts from Bosch Research, and the Carnegie Bosch Institute.

\bibliographystyle{acl_natbib}
\bibliography{acl2019}

\end{document}